\documentclass[11pt]{article}

\usepackage[final]{acl}

\usepackage{times}
\usepackage{latexsym}

\usepackage[T1]{fontenc}

\usepackage[utf8]{inputenc}

\usepackage{microtype}

\usepackage{inconsolata}

\usepackage{graphicx}

\usepackage{booktabs}
\usepackage{colortbl}
\usepackage{amssymb}
\usepackage{amsmath}

\usepackage{tcolorbox}
\usepackage{enumitem}
\usepackage{algorithm}
\usepackage{algorithmic}

\newtcolorbox{promptbox}[2][]{
  colback=white,            
  boxrule=0.8pt,            
  arc=2mm,                  
  fonttitle=\bfseries\sffamily, 
  fontupper=\small\ttfamily,    
  left=3mm, right=3mm, top=3mm, bottom=3mm, 
  title={#2},               
  #1                        
}

\newtcolorbox{examplebox}[2][]{
  colback=white,
  colframe=gray!75!black,
  colbacktitle=gray!15,
  coltitle=black,
  boxrule=0.5pt,
  arc=1mm,
  fonttitle=\bfseries\sffamily,
  fontupper=\small,
  left=3mm, right=3mm, top=3mm, bottom=3mm,
  title={#2},
  #1
}

\title{
PhageBench: Can LLMs Understand Raw Bacteriophage Genomes?
}

\author{First Author \\
Affiliation / Address line 1 \\
Affiliation / Address line 2 \\
Affiliation / Address line 3 \\
\texttt{email@domain} \\
\And Second Author \\
Affiliation / Address line 1 \\
Affiliation / Address line 2 \\
Affiliation / Address line 3 \\
\texttt{email@domain} \\}

\author{
 \textbf{Yusen Hou\textsuperscript{1,$\dagger$}},
 \textbf{Weicai Long\textsuperscript{1,$\dagger$}},
 \textbf{Haitao Hu\textsuperscript{1}},
 \textbf{Houcheng Su\textsuperscript{1}},
\\
 \textbf{Junning Feng\textsuperscript{1}},
 \textbf{Yanlin Zhang\textsuperscript{1,*}},
\\
 \textsuperscript{1}Hong Kong University of Science and Technology (Guangzhou), \\
 \small{
   \textbf{Correspondence*:} \href{yanlinzhang@hkust-gz.edu.cn}{yanlinzhang@hkust-gz.edu.cn}
 }
}

\begin{document}
\maketitle

\begingroup
\renewcommand\thefootnote{$\dagger$}
\footnotetext{Co-first authors}
\endgroup
\begin{abstract}
    Bacteriophages, often referred to as the dark matter of the biosphere, play a critical role in regulating microbial ecosystems and in antibiotic alternatives. Thus, accurate interpretation of their genomes holds significant scientific and practical value. While general-purpose Large Language Models (LLMs) excel at understanding biological texts, their ability to directly interpret raw nucleotide sequences and perform biological reasoning remains underexplored. To address this, we introduce PhageBench, the first benchmark designed to evaluate phage genome understanding by mirroring the workflow of bioinformatics experts. The dataset contains 5,600 high-quality samples covering five core tasks across three stages: Screening, Quality Control, and Phenotype Annotation. Our evaluation of eight LLMs reveals that general-purpose reasoning models significantly outperform random baselines in phage contig identification and host prediction, demonstrating promising potential for genomic understanding. However, they exhibit significant limitations in complex reasoning tasks involving long-range dependencies and fine-grained functional localization. These findings highlight the necessity of developing next-generation models with enhanced reasoning capabilities for biological sequences\footnote{Our PhageBench dataset and evaluation code are available at https://anonymous.4open.science/r/PhageBench-D428}.
\end{abstract}

\begin{figure}[h]
    \centering
    \includegraphics[width=1\linewidth]{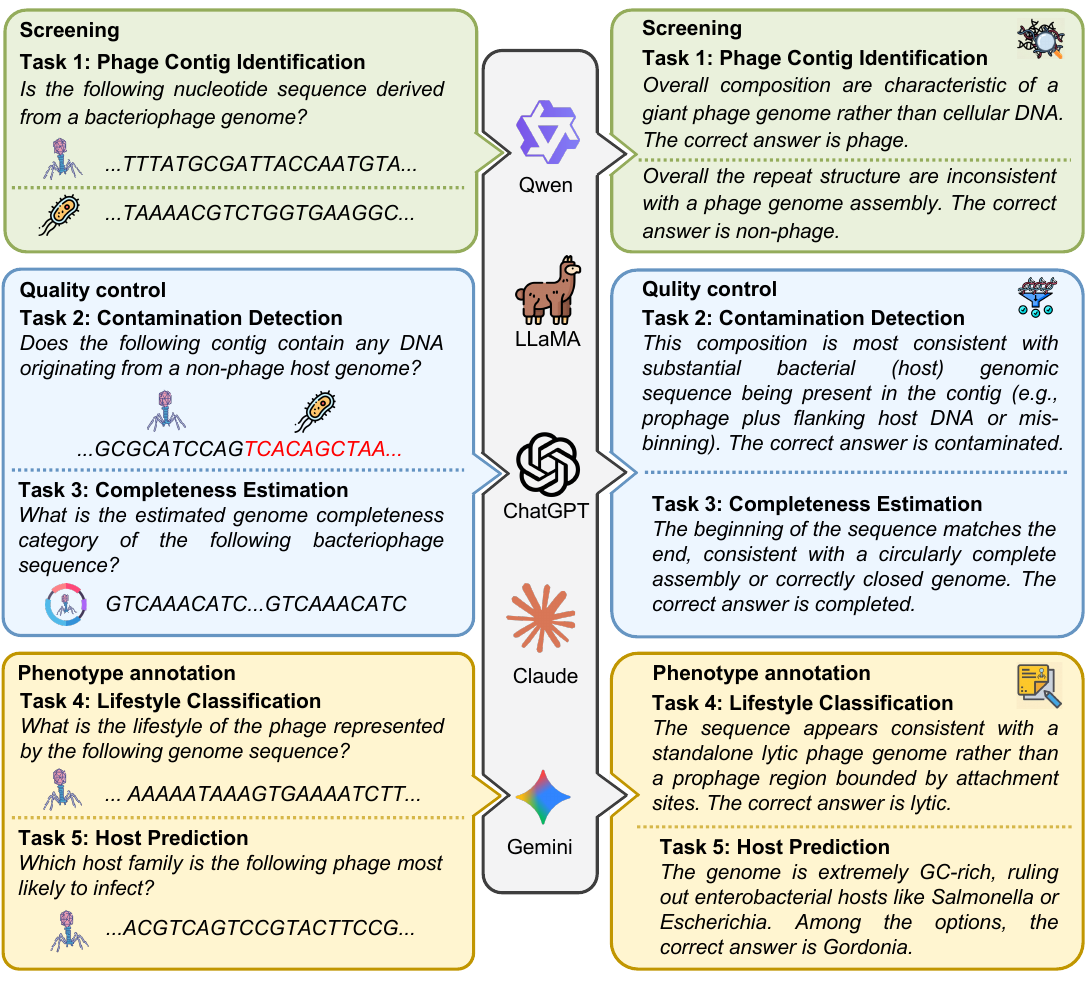}
    \caption{
        PhageBench challenges general-purpose LLMs to perform five phage analysis tasks, processing raw nucleotide sequence inputs to generate reasoning-based phenotypic predictions.    
    }
    \label{fig:overview}
\end{figure}

\section{Introduction}
The intersection of artificial intelligence and biology has shifted from analyzing biological texts to decoding, the language of life, DNA sequences. Recent Genomic Foundation Models (GFMs) such as, Evo \citep{nguyen2024sequence, brixi2025genome} and Caduceus\citep{schiff2024caduceus}, have demonstrated that deep learning architectures can capture the complex syntax and semantics of DNA. These models can predict gene expression~\citep{hou2024using,avsec2025alphagenome} and even design whole genome of functional bacteriophages (phages)~\citep{king2025generative}. However, these specialized models require extensive pre-training on DNA sequence corpora, similar to how general LLMs are trained on large text corpora. In parallel, general-purpose Large Language Models (LLMs) have exhibited strong reasoning abilities and performance on various biology-related tasks~\citep{qu2025crispr,swanson2025virtual,wang2025geneagent}. This raises a critical question: \textit{\textbf{Can general-purpose LLMs understand raw DNA sequences without specific pre-training?}}

To address this question, we propose a new challenge for the natural language processing (NLP) community: phage genome annotation. We select phages, viruses that infect bacteria, as the testbed for three reasons. First, the length of a typical phage genome ranges from approximately 3 to 150 kilobase pairs (kb)~\cite{dion2020phage}, which aligns well with the context windows of modern LLMs. This makes phages an ideal subject for evaluating non-natural language long-context reasoning. Second, unlike eukaryotic genomes which contain long non-coding introns, phage genomes exhibit a dense, linear organization where genes are arranged in a structured syntax similar to natural language~\cite{mavrich2017bacteriophage}. Previous studies indicate that this structural conservation allows for functional prediction based on genomic position even when sequence homology is low~\citep{grigson2025synteny}. This allows us to test whether LLMs can infer function in the absence of explicit sequence matches. Third, phages are the most abundant biological entities in the biosphere and play critical roles in regulating microbial ecosystems~\citep{rohwer2003global}, food engineering~\citep{prasad2025bacteriophage}, and phage therapy as alternatives to antibiotics~\citep{strathdee2023phage,skurnik2025phage}. However, the vast majority of phage sequences identified in metagenomic data remain unannotated dark matter lacking basic taxonomic and phenotypic information. This annotation bottleneck limits our ability to harness phages for therapeutic and biotechnological applications.

Therefore, achieving efficient and accurate phage genome annotation is of significant scientific and practical importance.
To systematically investigate the ability of LLMs to understand phage genomes, we introduce PhageBench, a multi-task benchmark designed to mirror the complete workflow of expert phage analysis. PhageBench follows a standard structure across three stages: Screening (phage contigs identification from environmental noise), Quality Control (contamination detection and completeness estimation), and Phenotype Annotation (lifestyle prediction and host prediction) (Fig. \ref{fig:overview}). This design reflects how biologists progressively filter and characterize sequences, moving from simple identification to complex biological inference. The benchmark contains over 5,600 samples across five tasks, with careful controls for sequence length distribution and class balance to ensure models cannot rely on statistical biases.

We evaluate eight advanced LLMs on PhageBench, including reasoning models. Our analysis reveals that general-purpose reasoning LLMs demonstrate promising potential in phage identification and host prediction tasks without domain-specific pre-training. However, they exhibit limitations in modeling long-range dependencies, leading to declines in performance in completeness estimation tasks involving global structures as sequence length increases. Furthermore, although these models are capable of applying accurate biological logic during reasoning, they struggle to precisely identify functional gene fragments within raw nucleotide sequences spanning tens of thousands of base pairs.
To summarize, our contributions are:
\begin{itemize}
    \item We first introduce raw phage genome annotation as a challenging testbed for evaluating the reasoning capabilities of general-purpose LLMs. This task requires models to process long-context sequences and perform zero-shot logical reasoning, providing a proxy for measuring LLM capabilities in long-range and complex syntactic analysis in non-natural language domains.
    \item We construct and release PhageBench, a comprehensive dataset containing 5,600 high-quality samples. PhageBench mirrors the real workflow of bioinformatics experts, covering three critical stages in phage genome annotation.
    \item By applying general-purpose LLMs to raw phage genome annotation, PhageBench provides a novel analysis pathway for genomics that does not depend on sequence homology. Our evaluation demonstrates the potential of LLMs for understanding raw genomic sequences, offering preliminary evidence for the feasibility of using general reasoning engines to accelerate the discovery, annotation, and therapeutic development of unknown phages in metagenomic data.
\end{itemize}

\begin{figure*}[t]
    \includegraphics[width=\linewidth]{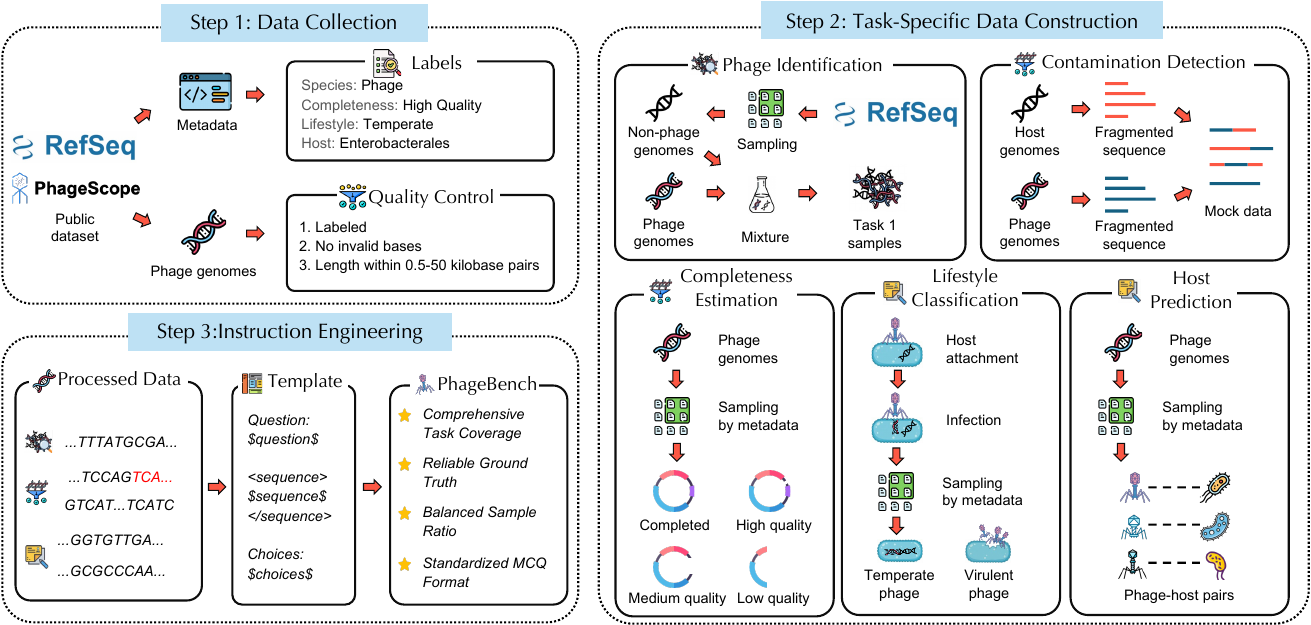}
    \caption{The overview of data construction of PhageBench. We filter raw sequences from public repositories, apply task-specific processing rules, and format the resulting samples into standardized instructions for evaluation. For detailed procedures for each task, please refer to Appendix \ref{sec:data-construction}.
    }
    \label{fig:data-construction}
\end{figure*}

\section{Related Work}
\subsection{Phage Genome Analysis}
The computational analysis of phage genomes has evolved from alignment-based sequence matching to feature-learning approaches. Traditional methods, such as BLAST~\cite{altschul1990basic}, rely on homology search against reference databases like GenBank~\citep{benson2018genbank}. While accurate for known sequences, these methods fail to identify sequences with low homology. To address this, learning-based approaches have been developed~\cite{camargo2024identification,peng2024viralm}. VirSorter2~\cite{guo2021virsorter2} integrates multiple classifiers to capture diverse viral signals, while CheckV~\cite{nayfach2021checkv} assesses genome completeness based on gene content and terminal repeats. However, these discriminative models typically output binary classifications without biological reasoning and often specialized for a single task. Our work investigates whether general-purpose LLMs can bridge this gap by leveraging their inherent reasoning capabilities to analyze raw genomic sequences directly.

\subsection{Genomic Foundation Models and Benchmarks}
The intersection of AI and genomics has spurred the development of specialized GFMs~\cite{zhou2023dnabert,nguyen2023hyenadna,dalla2025nucleotide}. Models such as DNABERT~\cite{ji2021dnabert} apply Transformer architectures to DNA sequences. Recent generative models, such as Evo, have demonstrated the ability to design whole phage genomes~\cite{king2025generative}. However, these specialized models require resource-intensive pre-training on massive genomic corpora. While some efforts, such as L2G~\cite{cheng2024l2g}, explore repurposing general LLMs for genomics, they still rely on extensive fine-tuning.

Existing genomic benchmarks, such as GenomicBenchmarks~\cite{grevsova2023genomic} and DNALongBench~\cite{cheng2025dnalongbench}, are primarily designed to evaluate GFMs and predominantly focus on mammalian genomic features, such as chromatin accessibility and complex regulatory mechanisms. These tasks differ significantly from the syntactic gene organization logic of prokaryotic phages. On the other hand, benchmarks designed for general LLMs, such as GeneTuring~\cite{shang2025benchmarking}, focus on knowledge retrieval rather than reasoning on raw genomic sequences. PhageBench prioritizes the understanding and reasoning of raw phage genome sequences. To the best of our knowledge, this is the first study to evaluate general-purpose LLMs on authentic whole-genome analysis tasks, thereby deepening our understanding of the capabilities of LLMs within the life sciences.

\section{The PhageBench Dataset}
\subsection{Underlying Principles}
While existing benchmarks in biology primarily focus on natural language-based tasks, such as question answering, information retrieval, or text summarization, PhageBench establishes a comprehensive evaluation dataset that integrates biological reasoning with the direct understanding and annotation of raw phage genomic sequences. Aiming to provide a rigorous standard for assessing off-the-shelf models, PhageBench examines the potential of general-purpose NLP technologies in driving scientific discovery within biology.

\begin{figure}[ht]
    \centering
    \includegraphics[width=0.99\linewidth]{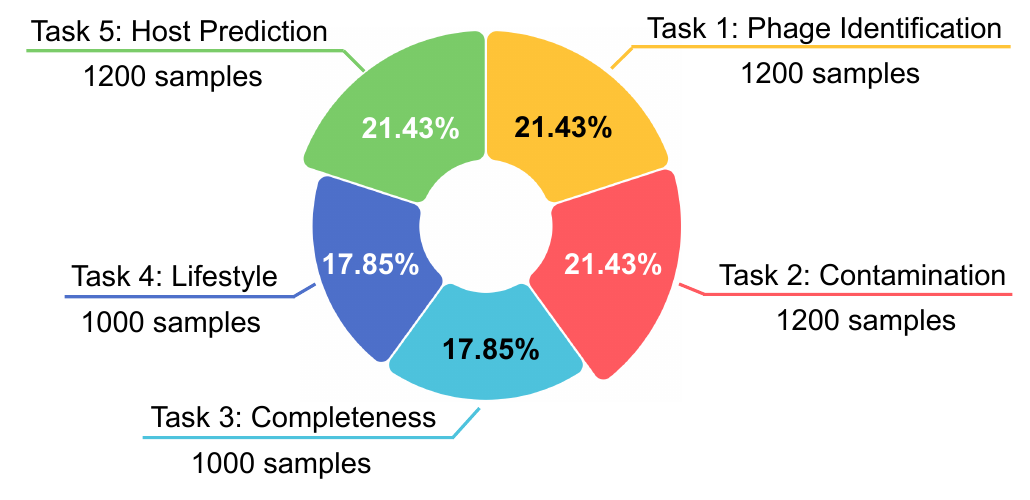}
    \caption{The compositional structure of PhageBench. The sizes of the constructed instruction datasets are labeled below the horizontal lines, and the percentages on the pie charts represent the proportion of data within each major category.}
    \label{fig:data-overview}
\end{figure}

The construction of PhageBench is guided by four core design principles: (1) encompassing three distinct analysis stages mapped into five core tasks to cover the full spectrum of phage genome analysis; (2) ensuring all sequences are derived from authoritative databases to serve as high-confidence ground truth; (3) controlling the ratio of positive to negative samples; and (4) adopting a standardized multiple-choice question format.

\subsection{Tasks Construction}

Phage genome analysis requires identifying entities from complex background sequences, assessing their quality, and ultimately inferring their biological functions. PhageBench mirrors this path through three distinct stages: Screening, Quality Control, and Phenotype Annotation. The overall data construction pipeline is illustrated in Fig.~\ref{fig:data-construction}. We collected raw sequences and metadata from authoritative databases, applied rigorous task-specific processing rules, and used only complete-level phage genomes for all tasks except completeness estimation task. Detailed construction procedures are provided in Appendix~\ref{sec:data-construction}.

\textbf{Stage 1: Screening.}
Metagenomic sequencing has revolutionized our access to total genetic material, yet it inherently yields a complex admixture of sequences derived from bacteria, archaea, eukaryotes, and viruses. Consequently, the foundational step in phage annotation is the accurate extraction of phage signals from this multispecies background. 

\textit{Task 1: Phage Contig Identification} functions as a binary classification challenge designed to distinguish phage sequences from a heterogeneous array of non-phage genomic DNA. This task evaluates the capacity of a model to discriminate specific phage signatures from diverse biological confounders, including non-phage viruses, protozoa, plasmids, fungi, bacteria, and archaea.

\textbf{Stage 2: Quality Control.}
Even within identified phage contigs, sequencing and assembly processes may yield fragmented genomes lacking essential genomic regions or chimeric sequence containing host DNA. These quality issues can mislead downstream analyses. Therefore, this stage evaluates the capacity of LLMs to assess both sequence purity and structural integrity.

\textit{Task 2: Contamination Detection} focuses on the identification of host genomic fragments that have been erroneously assembled with phage sequences. Unlike the previous screening stage, this task operates on the premise that the input sequences are phage-associated and challenges the model to discern whether a given contig is pure or contaminated by host-derived genetic sequences.

\textit{Task 3: Completeness Estimation} requires the model to evaluate the assembly quality of a phage genome, addressing the prevalence of partial viral sequences in metagenomic data due to insufficient sequencing depth. This task is formulated as a four-way classification problem where the model must categorize sequences into \textsc{Complete} (representing full-length genomes), \textsc{High-quality}, \textsc{Medium-quality}, and \textsc{Low-quality} tiers based on their genomic integrity and the presence of features such as terminal repeats.

\begin{figure*}[ht]
    \centering
    \includegraphics[width=0.99\linewidth]{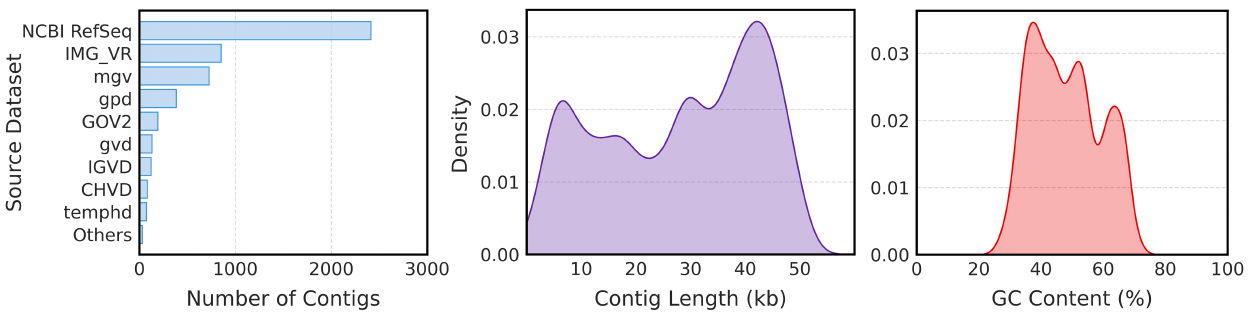}
    \caption{Statistical characteristics of the genomic sequences within the PhageBench dataset. The left panel displays the distribution of sequence sources. The central panel illustrates the probability density of contig lengths. The right panel depicts the distribution of GC content.}
    \label{fig:data-stats}
\end{figure*}

\textbf{Stage 3: Phenotype Annotation.} The final stage demands high-level biological reasoning to infer functional traits and ecological roles directly from genomic sequences. Distinguishing between virulent phages, which drive the lytic cycle, and temperate phages, capable of genomic integration, is essential for applications ranging from antibiotic alternatives to genome engineering. Furthermore, accurate host prediction remains a cornerstone for the effective deployment of phage therapy and biocontrol strategies.

\textit{Task 4: Lifestyle Classification} requires the model to predict the survival strategy of a phage by distinguishing between virulent and temperate lifestyles. This binary classification task evaluates the ability of an LLM to recognize specific genomic signatures associated with integration machinery, such as integrases and attachment sites, versus those indicative of exclusive lytic cycles.

\textit{Task 5: Host Prediction} challenges the model to identify the specific bacterial host of a phage across three taxonomic levels, including Order, Family, and Genus. This multi-class classification problem requires the decoding of subtle interaction signals, such as codon usage bias or receptor-binding protein domains, to infer ecological relationships directly from raw genomic sequences without reliance on sequence homology.

\textbf{Instruction Engineering.}
To better evaluate the ability of LLMs to understand and annotate raw phage genomic sequences, we designed task-specific instructions for each task category. To reduce data bias and ensure evaluation balance, all samples are formatted as multiple-choice questions: binary classification tasks (Tasks 1, 2, and 4) use two options, while multi-class classification tasks (Tasks 3 and 5) use four options. We randomly shuffled the option order for each sample to prevent position bias. 

\subsection{Dataset Statistics}

Fig.~\ref{fig:data-overview} and Fig.~\ref{fig:data-stats} summarize the statistical properties of tasks in PhageBench. The dataset comprises a total of 5,600 samples across five tasks: Tasks 1, 2, and 5 each contain 1,200 samples, while Tasks 3 and 4 each contain 1,000 samples. All phage genomic sequences range from 500 bp to 50 kb in length, with a mean length of 28.15 kb and a median of 29.82 kb. The GC content across all phage sequences has a mean of 48.06\% and a median of 47.00\%, reflecting typical phage genomic composition. These statistics, along with the diverse distribution of sequence sources illustrated in Fig.~\ref{fig:data-stats}, demonstrate that PhageBench encompasses a representative collection of phage genomic sequences suitable for evaluating LLM reasoning capabilities. More detailed statistics are provided in Appendix~\ref{sec:detailed_data_stats}.

\begin{table*}[!htbp]
\centering
\caption{Model performance accuracy (\%) on PhageBench tasks under zero shot CoT setting. Bold: the best performance. Underlined: the second performance. Avg.: the average across all tasks. T: task.}
\resizebox{\textwidth}{!}{%
\begin{tabular}{lccccccc}
\toprule
\textbf{Model} & \textbf{\begin{tabular}[c]{@{}c@{}}T1: Phage\\Identification\end{tabular}} & \textbf{\begin{tabular}[c]{@{}c@{}}T2: Contamination\end{tabular}} & \textbf{\begin{tabular}[c]{@{}c@{}}T3: Completeness\end{tabular}} & \textbf{\begin{tabular}[c]{@{}c@{}}T4: Lifestyle\end{tabular}} & \textbf{\begin{tabular}[c]{@{}c@{}}T5: Host\\Prediction\end{tabular}} & \textbf{\begin{tabular}[c]{@{}c@{}}Avg.\end{tabular}} \\ \midrule
\multicolumn{7}{c}{\cellcolor[HTML]{E9EEEA}\textit{Non-reasoning}} \\ \midrule
GPT-4o-mini & 50.00 & 50.08 & 24.50 & 48.40 & 27.75 & 40.15 \\
LLaMA 4 & 53.25 & 49.58 & 27.10 & \underline{54.50} & 28.58 & 42.60 \\
Qwen3-235b & 51.58 & 49.33 & 25.40 & 52.00 & 28.58 & 41.38 \\ \midrule
\multicolumn{7}{c}{\cellcolor[HTML]{E9EEEA}\textit{Reasoning}} \\ \midrule
GPT-OSS-120b & 50.08 & 46.58 & 24.20 & 49.50 & 41.00 & 42.27 \\
GPT-5.2 & \underline{70.75} & \underline{54.17} & \textbf{43.10} & 51.50 & 35.25 & \underline{50.95} \\
Gemini-3-flash & \textbf{70.83} & \textbf{58.83} & 32.10 & \textbf{57.40} & \textbf{62.50} & \textbf{56.33} \\
Claude-sonnet-4.5 & 57.92 & 51.83 & \underline{35.50} & 50.00 & \underline{44.17} & 47.88 \\
Qwen3-max & 57.42 & 52.58 & 33.80 & 49.90 & 34.75 & 45.69 \\ \midrule
Random & 50.00 & 50.00 & 25.00 & 50.00 & 25.00 & 40.00 \\
\bottomrule
\end{tabular}%
}
\label{tab:phagebench_acc}
\end{table*}

\section{Evaluation}
We evaluate eight advanced LLMs spanning four categories based on weight accessibility and reasoning capability. \textit{Open-weight, non-reasoning} models include LLaMA-4~\citep{llama4} and Qwen3-235b~\citep{yang2025qwen3}. \textit{Closed-weight, non-reasoning} models include GPT-4o-mini~\citep{gpt-4o-mini}. \textit{Open-weight, reasoning} models include GPT-OSS-120b~\citep{agarwal2025gpt}. \textit{Closed-weight, reasoning} models include GPT-5.2~\citep{gpt52}, Gemini-3-flash~\citep{gemini-3-flash}, Claude-sonnet-4.5~\citep{claude_sonnet_4_5}, and Qwen3-max~\citep{qwen3_max}. This selection covers a diverse range of model architectures and capabilities, enabling comprehensive evaluation of LLMs' performance on raw phage genome understanding and analysis.

All tasks are evaluated under a zero-shot chain of thought setting without in-context examples, allowing us to assess the models' inherent biological sequence reasoning capabilities in realistic usage scenarios. For models with built-in reasoning capabilities, we enable these features by default.

We use classification accuracy as the primary evaluation metric. Since PhageBench maintains strict class balance across all tasks, accuracy provides an unbiased measure of model performance. Additional details on experimental settings are provided in Appendix~\ref{sec:evaluation_details}.
\subsection{LLMs' Performance on Raw Phage Genome Understanding}

Table~\ref{tab:phagebench_acc} presents the comprehensive performance of evaluated models across the five PhageBench tasks. The results indicate that while LLMs possess promising capabilities for decoding raw phage sequences, this domain remains a formidable challenge. Gemini-3-flash achieves the highest average accuracy at 56.33\%, significantly surpassing the random baseline, followed by GPT-5.2. In contrast, non-reasoning models largely cluster around the random baseline. We observe two consistent trends. First, reasoning-enhanced models outperform their non-reasoning counterparts, suggesting that the ability to generate intermediate logical steps facilitates phage genome comprehension. Second, closed-weight models maintain a distinct performance advantage over open-weight alternatives, highlighting a capability gap that remains to be bridged in raw phage genome understanding.

\textbf{Proficiency in Recognizing Statistical Genomic Features.}
Model performance is strongest when tasks rely on distinguishing broad compositional signatures. In phage contig identification task, Gemini-3-flash and GPT-5.2 achieve approximately 70\% accuracy. Similarly, in host prediction, Gemini-3-flash reaches 62.50\%, more than double the random baseline. These results imply that current LLMs can successfully discern the characteristic biases, such as codon usage or nucleotide composition, that differentiate phages from background noise or define host compatibility. This suggests models effectively treat DNA as a structured language with recognizable local syntax.

\textbf{Competence in Global Structural Assessment.}
Moving beyond statistical features, completeness estimation evaluates the capacity to process genomic information at a global scale. Top models reliably exceed the 25\% random baseline, with GPT-5.2 reaching 43.10\%, indicating a fundamental grasp of structural integrity. However, performance remains lower than in identification tasks. This intermediate performance likely reflects the challenge of long-range dependency, as determining completeness necessitates verifying distal structural features, such as direct terminal repeats (DTR) at the sequence boundaries. While LLMs exhibit awareness of these global structures, maintaining such extended coherence remains a bottleneck compared to statistical pattern recognizing.

\begin{figure*}[ht]
    \centering
    \includegraphics[width=1\linewidth]{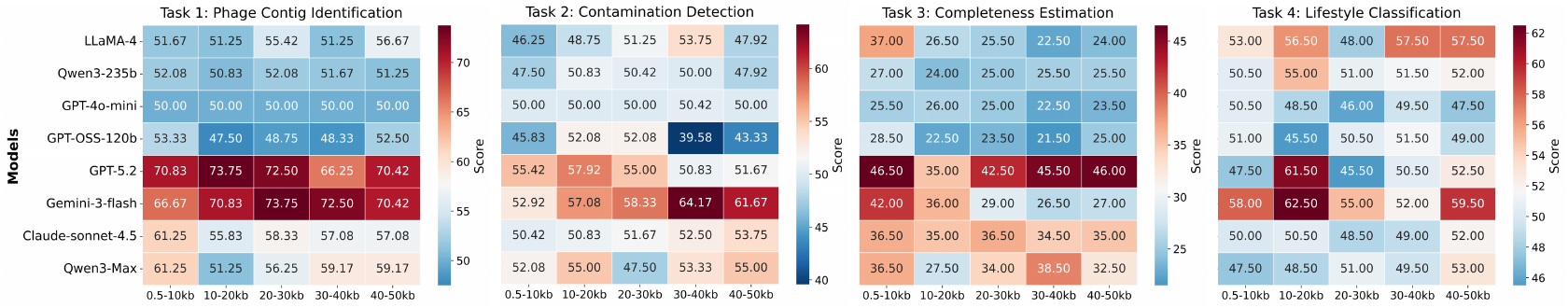}
    \caption{Performance heatmaps across sequence length bins for Tasks 1--4. Each cell represents the accuracy of a specific model on sequences within a particular length range.}
    \label{fig:results-length}
\end{figure*}

\begin{table*}[]
\centering
\caption{Comparison of model accuracy (\%) between zero-shot and zero-shot with CoT modes. Grey cells indicate the $\Delta$ = $Acc_{CoT}-Acc_{0-shot}$, which is the improvement of CoT. Positive values indicate CoT improvement. Bold values denote the best 0-shot performance per task.}
\resizebox{\textwidth}{!}{%
\begin{tabular}{lcccccccccccc}
\toprule
\textbf{Model} & \multicolumn{2}{c}{\textbf{\begin{tabular}[c]{@{}c@{}}T1: Phage\\Identification\end{tabular}}} & \multicolumn{2}{c}{\textbf{\begin{tabular}[c]{@{}c@{}}T2: Contamination\end{tabular}}} & \multicolumn{2}{c}{\textbf{\begin{tabular}[c]{@{}c@{}}T3: Completeness\end{tabular}}} & \multicolumn{2}{c}{\textbf{\begin{tabular}[c]{@{}c@{}}T4: Lifestyle\end{tabular}}} & \multicolumn{2}{c}{\textbf{\begin{tabular}[c]{@{}c@{}}T5: Host\\Prediction\end{tabular}}} & \multicolumn{2}{c}{\textbf{Average}} \\
 & \textbf{0-shot} & \cellcolor[HTML]{F0F0F0}\textbf{$\Delta$} & \textbf{0-shot} & \cellcolor[HTML]{F0F0F0}\textbf{$\Delta$} & \textbf{0-shot} & \cellcolor[HTML]{F0F0F0}\textbf{$\Delta$} & \textbf{0-shot} & \cellcolor[HTML]{F0F0F0}\textbf{$\Delta$} & \textbf{0-shot} & \cellcolor[HTML]{F0F0F0}\textbf{$\Delta$} & \textbf{0-shot} & \cellcolor[HTML]{F0F0F0}\textbf{$\Delta$} \\ \midrule
GPT-4o-mini & 50.00 & \cellcolor[HTML]{F0F0F0}0.00 & 50.00 & \cellcolor[HTML]{F0F0F0}+0.08 & 27.90 & \cellcolor[HTML]{F0F0F0}$-$3.40 & 48.20 & \cellcolor[HTML]{F0F0F0}+0.20 & 26.75 & \cellcolor[HTML]{F0F0F0}+1.00 & 40.57 & \cellcolor[HTML]{F0F0F0}$-$0.42 \\
LLaMA-4 & 51.75 & \cellcolor[HTML]{F0F0F0}+1.50 & 51.33 & \cellcolor[HTML]{F0F0F0}$-$1.75 & 25.60 & \cellcolor[HTML]{F0F0F0}+1.50 & 50.60 & \cellcolor[HTML]{F0F0F0}+3.90 & 27.83 & \cellcolor[HTML]{F0F0F0}+0.75 & 41.42 & \cellcolor[HTML]{F0F0F0}+1.18 \\
Qwen3-235b & 50.00 & \cellcolor[HTML]{F0F0F0}+1.58 & 50.42 & \cellcolor[HTML]{F0F0F0}$-$1.09 & 25.00 & \cellcolor[HTML]{F0F0F0}+0.40 & 50.60 & \cellcolor[HTML]{F0F0F0}+1.40 & 28.42 & \cellcolor[HTML]{F0F0F0}+0.16 & 40.89 & \cellcolor[HTML]{F0F0F0}+0.49 \\
GPT-5.2 & 68.50 & \cellcolor[HTML]{F0F0F0}+2.25 & 51.42 & \cellcolor[HTML]{F0F0F0}+2.75 & \textbf{36.60} & \cellcolor[HTML]{F0F0F0}+6.50 & 50.50 & \cellcolor[HTML]{F0F0F0}+1.00 & 29.17 & \cellcolor[HTML]{F0F0F0}+6.08 & 47.24 & \cellcolor[HTML]{F0F0F0}+3.71 \\
Gemini-3-flash & \textbf{70.25} & \cellcolor[HTML]{F0F0F0}+0.58 & \textbf{56.42} & \cellcolor[HTML]{F0F0F0}+2.41 & 30.20 & \cellcolor[HTML]{F0F0F0}+1.90 & \textbf{58.20} & \cellcolor[HTML]{F0F0F0}$-$0.80 & \textbf{64.58} & \cellcolor[HTML]{F0F0F0}$-$2.08 & \textbf{55.93} & \cellcolor[HTML]{F0F0F0}+0.40 \\
Qwen3-Max & 50.08 & \cellcolor[HTML]{F0F0F0}+7.34 & 50.33 & \cellcolor[HTML]{F0F0F0}+2.25 & 24.90 & \cellcolor[HTML]{F0F0F0}+8.90 & 48.70 & \cellcolor[HTML]{F0F0F0}+1.20 & 27.58 & \cellcolor[HTML]{F0F0F0}+7.17 & 40.32 & \cellcolor[HTML]{F0F0F0}+5.37 \\
\bottomrule
\end{tabular}%
}
\label{tab:cot_comparison}
\end{table*}

\textbf{Challenges in Fine-grained Inference.}
Conversely, models encounter significant obstacles when tasks demand disentangling highly similar sequences or identifying sparse functional motifs. Contamination detection proves difficult because phage and host sequences share high statistical similarity driven by long-term coevolution, rendering simple statistical features matching insufficient. Lifestyle classification presents a profound challenge in functional inference. Unlike taxonomic classification, differentiating virulent from temperate phages requires identifying specific elements, such as integrases, within long sequences. Current models struggle to perform this level of precise retrieval and reasoning directly from raw nucleotides.

\subsection{Performance Across Sequence Length}
Phage genomes exhibit significant diversity in length. This biological variation raises a critical question for practical application: \textit{at what genome scale can LLMs reliably interpret raw genomic sequences?} To answer this, we analyzed model accuracy across length bins ranging from 0.5kb to 50kb for Tasks 1--4 (Fig.~\ref{fig:results-length}), excluding Task 5 due to sampling limitations. 

The results reveal distinct behavioral patterns rooted in the nature of each task. For phage identification, top models display remarkable stability across all length bins. This length-invariance suggests that the discriminative features are syntactic in nature and remain easily detectable regardless of context size. Conversely, completeness estimation exposes the limitations of global structural processing, where accuracy declines notably as sequences approach 50kb. This degradation reflects a failure in long-range dependency, as current architectures struggle to link distal features required to verify genome closure in extended contexts.

Complex inference tasks exhibit divergent responses to increased context. Contamination detection benefits from longer sequences, as extended inputs allow for evidence accumulation of heterogeneous signals that reveal host-derived fragments. In contrast, lifestyle classification remains near-random across all lengths. This indicates a semantic bottleneck rather than a contextual one, implying that the primary challenge lies in the precise retrieval of sparse functional markers which is not ameliorated simply by providing additional nucleotide context.

\subsection{Impact of Test-Time Scaling}

While our main evaluation shown in Table~\ref{tab:phagebench_acc} utilizes CoT prompting to maximize model capabilities, recent studies suggest that explicit reasoning is not universally beneficial across all domains~\citep{wang2023towards,turpin2023language}. To quantify the extent to which test-time computational scaling specifically enhances raw genomic understanding, we conducted an ablation study comparing CoT against a direct zero-shot baseline.

As shown in Table~\ref{tab:cot_comparison}, allocating additional test-time computation generally serves as a powerful amplifier for decoding phage genomes. The majority of models exhibit positive performance gains when reasoning is activated. This confirms that for biological sequences, thinking time can effectively translate into performance gains.

However, we observe distinct performance regressions in specific LLMs, such as Gemini-3-flash on host prediction. In these specific cases, the model's intuitive alignment with genomic patterns proves more robust than its generated reasoning chains, highlighting the nuance required when applying test-time scaling to biological data.

\begin{figure}[ht]
    \centering
    \includegraphics[width=0.88\linewidth]{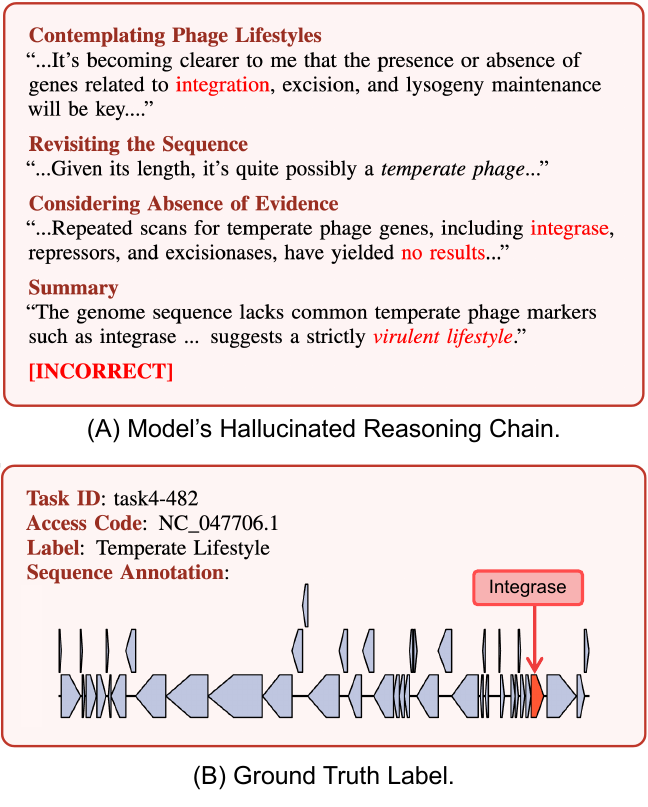}
    \caption{A case study of hallucinated reasoning in Lifestyle Classification. (A) While the model correctly identifies integrase as a key marker, it incorrectly asserts its absence in the sequence.(B) The ground truth annotation confirms the presence of integrase.}
    \label{fig:case-example}
\end{figure}

\subsection{Success and Failure Analysis}
\label{sec:success_failure}

To understand the factors underlying model performance, we analyzed their accuracy stratified by ground-truth labels (see Appendix~\ref{sec:deep-ana}), and examined the chain-of-thought reasoning traces of Gemini-3-flash.

Our analysis reveals that leading models possess genuine pattern recognition capabilities for genomic sequences. In phage identification, Gemini-3-flash achieves 99.0\% accuracy on true phages. Crucially, its performance on non-phage samples tracks with biological similarity: accuracy is high for eukaryotic protozoa (69.0\%) but drops to 10.0\% for prokaryotic plasmids, which share genomic features with phages. The reasoning traces frequently cite nucleotide composition, GC content distributions, and the presence of open reading frames as discriminative features. This implies the model successfully discriminates broad biological domains but struggles with fine-grained structural mimics.

However, these models are fundamentally constrained by a disconnect between abstract biological knowledge and precise sequence grounding. The most common errors made by Gemini-3-flash involve explicit claims about gene absence or existence contrary to fact. As visualized in Fig.~\ref{fig:case-example}, the model generates biologically plausible reasoning without the capacity to verify specific claims against the input sequence. Further qualitative analysis is provided in Appendix~\ref{sec:qualitative_results}.

\section{Findings and Open Directions}

While general-purpose LLMs demonstrate promising capabilities in genomic interpretation, our evaluation uncovers significant deficiencies in fine-grained sequence reasoning and long-range dependency modeling. To accelerate scientific discovery in phage genomics, we call for closer collaboration between the NLP and Biology communities. We highlight two critical directions for future research.

\textbf{Enhancing Sequence-Grounded Reasoning Capabilities.}
Our analysis suggests a fundamental disconnect between the model's abstract biological knowledge and its perception of raw nucleotide sequences. To bridge this gap, we propose two strategies. The first is Multi-modal Alignment. By fine-tuning LLMs on paired raw genomes and functional annotations, models can learn to map sequence motifs directly to biological phenotypes, moving beyond simple statistical correlations to semantic understanding. The second strategy involves Tool-Augmented Reasoning. Recognizing that visual scanning of long sequences is an inherent weakness of token-based models, future frameworks may adopt an \textit{Extract-then-Infer} workflow. By granting LLMs access to external tools, models can verify their intuitive hypotheses against rigorous computational results.

\textbf{Optimizing Global Dependency Modeling.}
The degradation of performance in completeness estimation tasks as sequence length increases highlights that LLMs struggle to maintain the global coherence necessary to correlate distal features, such as direct terminal repeats. Since determining genome completeness often relies on specific structural rules (e.g., checking if the 5' and 3' ends match), prompts that explicitly instruct the model to first extract the terminal sequences and then perform a local comparison may effectively bypass the interference of intermediate context. 

\section{Conclusion}

This study introduces PhageBench, a new benchmark designed to advance the phage genome annotation. By establishing a rigorous evaluation protocol, we assessed advanced LLMs and confirmed their latent capabilities for phage genome analysis. However, significant performance bottlenecks remain before LLMs can effectively drive scientific discovery in phage genomics, particularly regarding the critical challenge of grounding abstract biological knowledge in concrete sequence evidence. These findings offer novel insights for extending the utility of LLMs within the life sciences and establish a basis for future optimization. We hope that PhageBench will foster innovative collaboration between the NLP and life science communities, accelerating the discovery and application of phage dark matter in real-world scenarios.

\section*{Limitations}

Our study has three primary limitations regarding data scope, tokenization mechanisms, and model transparency. 

\textbf{Data}. To balance computational costs with current context window capabilities, PhageBench restricts genomic sequences to a maximum length of 50kb. While this covers a significant portion of the phage population, it inherently excludes larger phage entities, such as jumbo phages (genomes exceeding 200kb), which often harbor complex metabolic genes and distinct evolutionary strategies. Consequently, our findings may not fully generalize to these more extensive and complex genomic architectures.

\textbf{Tokenization}. We evaluated models using their default tokenization strategies, which utilize Byte Pair Encoding (BPE) optimized for natural language. This approach is likely suboptimal for genomic data, as BPE tends to aggregate nucleotide subsequences based on statistical frequency rather than biological significance, potentially obscuring functional motifs. Future work could explore alternative preprocessing strategies, such as inserting spaces to enforce single-base tokenization or adopting k-mer based aggregation, to better align input representations with biological logic.

\textbf{Models}. The inclusion of closed-source models introduces the unavoidable risk of data contamination. Since the pre-training corpora of these models are not public, we cannot definitively rule out the possibility that sequences from PhageBench were present in their training data. However, the frequent occurrence of hallucinatory reasoning observed in our error analysis suggests that even if models have encountered these sequences, they rely heavily on imperfect recall or probabilistic guessing rather than robust memorization.

\section*{Ethical Considerations}
The PhageBench dataset presented in this study is constructed entirely from publicly available genomic repositories. We confirm that no proprietary, confidential, or personally identifiable information is involved in this work, and all data collection procedures strictly adhere to established community norms and best practices for genomic data usage.

Regarding biosafety, phages are viruses that specifically infect and replicate within bacteria; they do not infect human cells and are generally considered to have a high safety profile. Furthermore, the primary objective of this benchmark is to evaluate the capability of LLMs to understand and reason over existing biological sequences. This study does not involve the generative design of novel biological entities or the modification of functional pathogens.

However, we acknowledge the broader ethical implications and potential dual-use risks associated with the intersection of artificial intelligence and biology. We urge all users utilizing PhageBench to strictly adhere to ethical guidelines and biosafety regulations governing computational biology to ensure the responsible development of AI technologies.

\bibliography{custom}

\appendix
\section{Broader Impact}
\label{sec:broader_impact}

This work extends beyond specific phage genome annotation tasks, holding significant implications for both the artificial intelligence and computational biology communities. By introducing PhageBench, we push the frontier of Large Language Models from processing scientific literature to decoding the fundamental language of life itself, highlighting the urgent need for models that adhere to strict biological rules rather than relying on probabilistic approximation. On a societal level, efficient genome annotation is critical for addressing the antibiotic resistance crisis through phage therapy, which is currently bottlenecked by uncharacterized phage dark matter. Our study explores the potential of LLMs to streamline the screening and characterization of these entities, offering a perspective on how AI might assist in shortening the timeline from environmental discovery to clinical application. Ultimately, we hope this benchmark serves as a catalyst for interdisciplinary collaboration, demystifying genomic data for AI researchers while objectively demonstrating the current capabilities and limitations of LLMs to the biological community.

\section{PhageBench Benchmark Details}
\label{sec:data_detail}

\subsection{Detailed Data Construction}
\label{sec:data-construction}

\subsubsection{Source Data Collection and Preprocessing}
The foundation of the PhageBench dataset is built upon high-quality genomic sequences derived from authoritative repositories. We utilized PhageScope \citep{wang2024phagescope} and NCBI RefSeq~\citep{o2016reference} as the primary reservoir for phage genomes. PhageScope is a comprehensive database that aggregates sequences from multiple public archives, including GenBank~\citep{benson2018genbank}, RefSeq~\citep{o2016reference}, PhagesDB~\citep{russell2017phagesdb}, GOV2~\citep{gregory2019marine}, GVD~\citep{gregory2020gut}, and MGV~\citep{nayfach2021metagenomic}, while providing systematic annotations generated by fifteen SOTA bioinformatics tools. This integration ensures that our positive samples represent a diverse and high-confidence collection of phage genomes. For the construction of negative samples required in phage contig identification task, we sourced non-phage sequences directly from the NCBI RefSeq database. We applied a strict length filtering criterion. Only sequences with clear labels and lengths ranging from 0.5 to 50 kilobases were retained. Furthermore, we performed a quality control step to exclude sequences containing invalid bases, such as ambiguous 'N' characters, ensuring that the model input consists solely of high-quality raw nucleotide sequences.

\subsubsection{Construction of Phage Contig Identification}
To evaluate the capability of LLMs in distinguishing phage sequences from environmental noise, we constructed a binary classification dataset with a balanced distribution of positive and negative samples. The negative samples were curated from the NCBI RefSeq database, spanning six distinct biological categories: non-phage viruses, protozoa, plasmids, fungi, bacteria, and archaea. We implemented a category-specific processing strategy to handle the significant length variations among these biological entities. For non-phage viruses, protozoa, and plasmids, we utilized their complete genomic sequences, as their natural lengths generally fall within our target distribution. In contrast, for fungi, bacteria, and archaea, which typically possess much larger genomes, we applied a random cropping strategy to generate fragmented sequences. This process ensured that the length distribution of the negative fragments strictly matched that of the positive phage samples. To prevent the model from distinguishing classes based solely on sequence length, we maintained a strict 1:1 ratio between phage and non-phage samples, and within the negative set, the six biological sources were also balanced equally to ensure diversity.

\subsubsection{Mock Data Generation for Contamination Detection}
Unlike random noise insertion, we followed previous work \citep{peng2025viralqc} to simulate realistic biological contamination events. We selected host fragments based on the ground-truth host metadata provided by PhageScope, ensuring that the contaminant sequences correspond to the actual biological hosts of the phages. We defined the contamination ratio ($r$) as the length of the host fragment divided by the total sequence length. To assess model sensitivity, we generated chimeric samples at three specific contamination ratios: 12.5\%, 25\%, and 50\%. We employed three distinct insertion strategies, Prefix, Suffix and Internal, to simulate different assembly error patterns as shown in Fig.~\ref{fig:data-construction}. The generation process is formally described in Algorithm \ref{alg:chimera}.

\begin{algorithm}[h]
\caption{Mock Data Generation}
\label{alg:chimera}
\begin{algorithmic}[1]
\REQUIRE Phage Sequence $S_p$, Host Sequence $S_h$, Contamination Ratio $r$, Mode $M$
\ENSURE Chimeric Sequence $S_{chimera}$
\STATE Calculate target total length $L_{total} \leftarrow Length(S_p) / (1-r)$
\STATE Calculate host fragment length $L_{host} \leftarrow L_{total} \times r$
\STATE Extract host fragment $F_h \leftarrow S_h[\text{random} : \text{random} + L_{host}]$
\IF{$M$ is PREFIX}
    \STATE $S_{chimera} \leftarrow F_h + S_p$
\ELSIF{$M$ is SUFFIX}
    \STATE $S_{chimera} \leftarrow S_p + F_h$
\ELSIF{$M$ is INTERNAL}
    \STATE Select random split point $k$ where $0 < k < Length(S_p)$
    \STATE $S_{chimera} \leftarrow S_p[0:k] + F_h + S_p[k:end]$
\ENDIF
\RETURN $S_{chimera}$
\end{algorithmic}
\end{algorithm}

\begin{table*}[t]
\centering
\caption{The specific instructions designed for each task in PhageBench.}
\label{tab:instructions}
\resizebox{0.95\textwidth}{!}{%
\begin{tabular}{lp{12cm}}
\toprule
\textbf{Task} & \textbf{Question Template} \\ \midrule
Phage Contig Identification & Is the following nucleotide sequence derived from a bacteriophage genome? \\
Contamination Classification & This sequence is a phage-associated contig. Does it contain any DNA originating from a non-phage host genome? \\
Completeness Classification & What is the estimated genome completeness category of the following bacteriophage sequence? \\
Lifestyle Classification & What is the most likely lifestyle of the bacteriophage represented by the following genome sequence? \\
Host Prediction & Which host family/genus/order is the following bacteriophage most likely to infect? \\ \bottomrule
\end{tabular}%
}
\end{table*}

\subsubsection{Construction of Completeness Estimation}
For genome completeness estimation, we extracted high-confidence completeness labels directly from the PhageScope database, which uses the standard tool CheckV~\citep{nayfach2021checkv} for quality assessment. The samples were categorized into four classes: Complete, High-quality, Medium-quality, and Low-quality. To ensure a rigorous evaluation, we applied strict length binning and class balancing. Specifically, we ensured an equal number of samples for each of the four completeness categories within every defined length interval. 

\subsubsection{Construction of Phenotype Annotation}
The phenotype annotation stage consists of lifestyle classification and host prediction. For lifestyle classification, we extracted lifestyle labels directly from PhageScope. For host prediction, we followed previous work \citep{shang2025genomic} to use the RefSeq Virus–Host Database (RefSeq-VHDB) as the data source. To construct the multiple-choice questions for this task, we implemented a hardness-aware distractor generation strategy. The incorrect options, or distractors, were randomly selected from different hosts within the same taxonomic level as the correct answer. For instance, if the ground truth host belongs to a specific Family, the distractors were sampled from other bacterial Families. This design challenges the model to distinguish between biologically comparable entities rather than identifying easy out-of-distribution targets.

\subsubsection{Instruction Design and MCQ Construction}
To transform the biological data into a format suitable for LLM evaluation, we employed an instruction engineering process. Each raw nucleotide sequence was encapsulated within a task-specific prompt template designed to simulate a bioinformatics query; the specific instructions and templates for each task are presented in Table~\ref{tab:instructions}. To standardize the evaluation and eliminate potential position bias, the order of the multiple-choice options was randomly shuffled for every sample. Finally, we required all models to generate outputs in a structured JSON format. This standardized output format facilitates automated parsing and ensures the accuracy of the quantitative evaluation.

\subsection{Detailed Dataset Statistics}
\label{sec:detailed_data_stats}

\subsubsection{General Statistical Characteristics}
The PhageBench dataset is constructed to ensure robust statistical properties that reflect real-world biological diversity while maintaining computational feasibility for LLMs. As illustrated in Fig.~\ref{fig:data-stats}, the dataset aggregates genomic sequences from a wide array of sources, ranging from environmental metagenomes found in databases such as GOV2 and GVD to high-quality isolates from NCBI RefSeq. This diversity ensures that LLMs are evaluated across various ecological niches and sequencing protocols. Regarding sequence properties, we strictly controlled the length of genomic segments to fall within the range of 0.5 to 50 kb. The resulting dataset exhibits a mean sequence length of approximately 28.15 kb and a median of 29.82 kb. Furthermore, the GC content of the sequences spans a wide dynamic range with a mean of 48.06\%, indicating that the dataset covers phages adapted to diverse host environments.

\begin{figure*}[ht]
    \centering
    \includegraphics[width=0.99\linewidth]{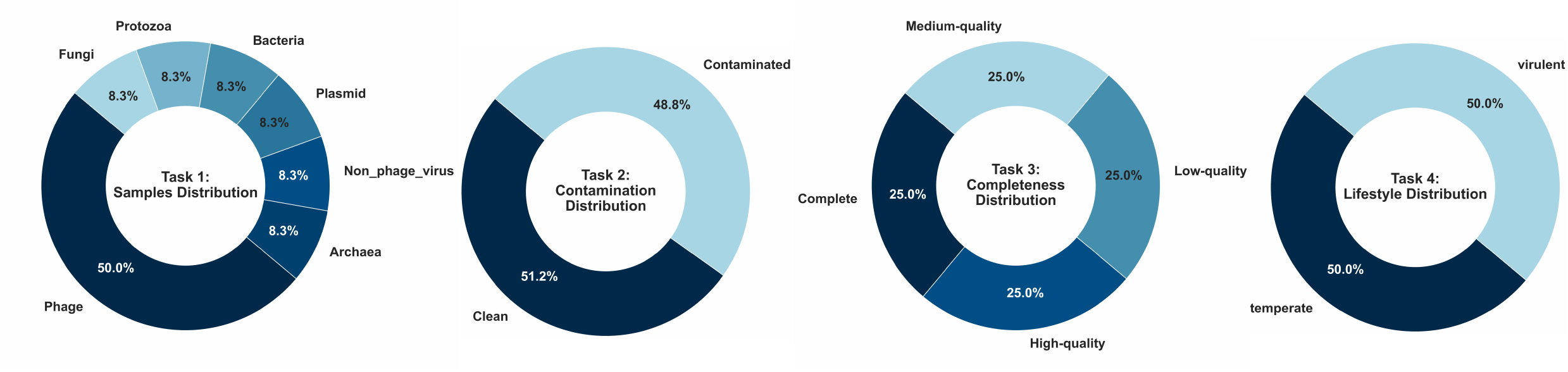}
    \caption{Detailed class and subclass distributions across the task 1--4. The pie charts demonstrate the strict balancing strategy applied to primary classes and quality tiers ensuring statistical fairness.}
    \label{fig:data-detailed}
\end{figure*}

\begin{figure*}[ht]
    \centering
    \includegraphics[width=0.99\linewidth]{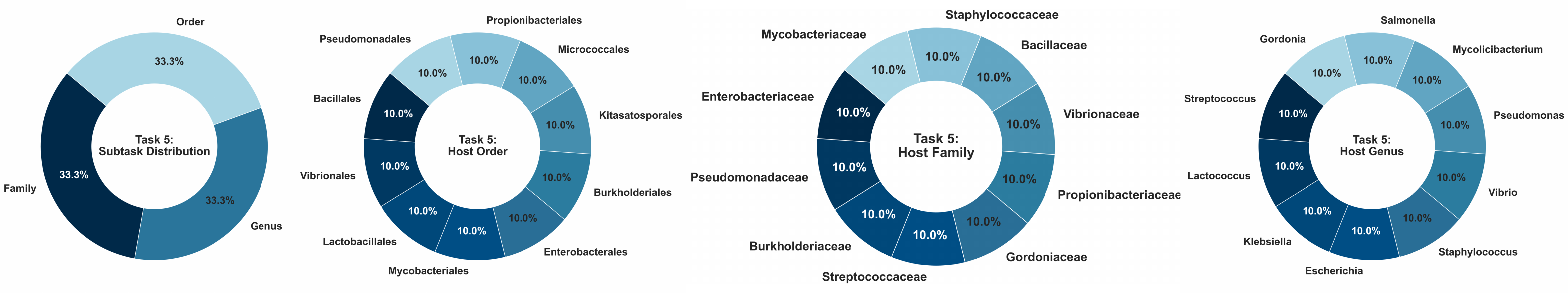}
    \caption{Detailed class distributions of task 5. The pie charts demonstrate the strict balancing strategy applied to fine-grained taxonomic categories.}
    \label{fig:data-detailed2}
\end{figure*}

\begin{figure}[h]
    \centering
    \includegraphics[width=0.95\linewidth]{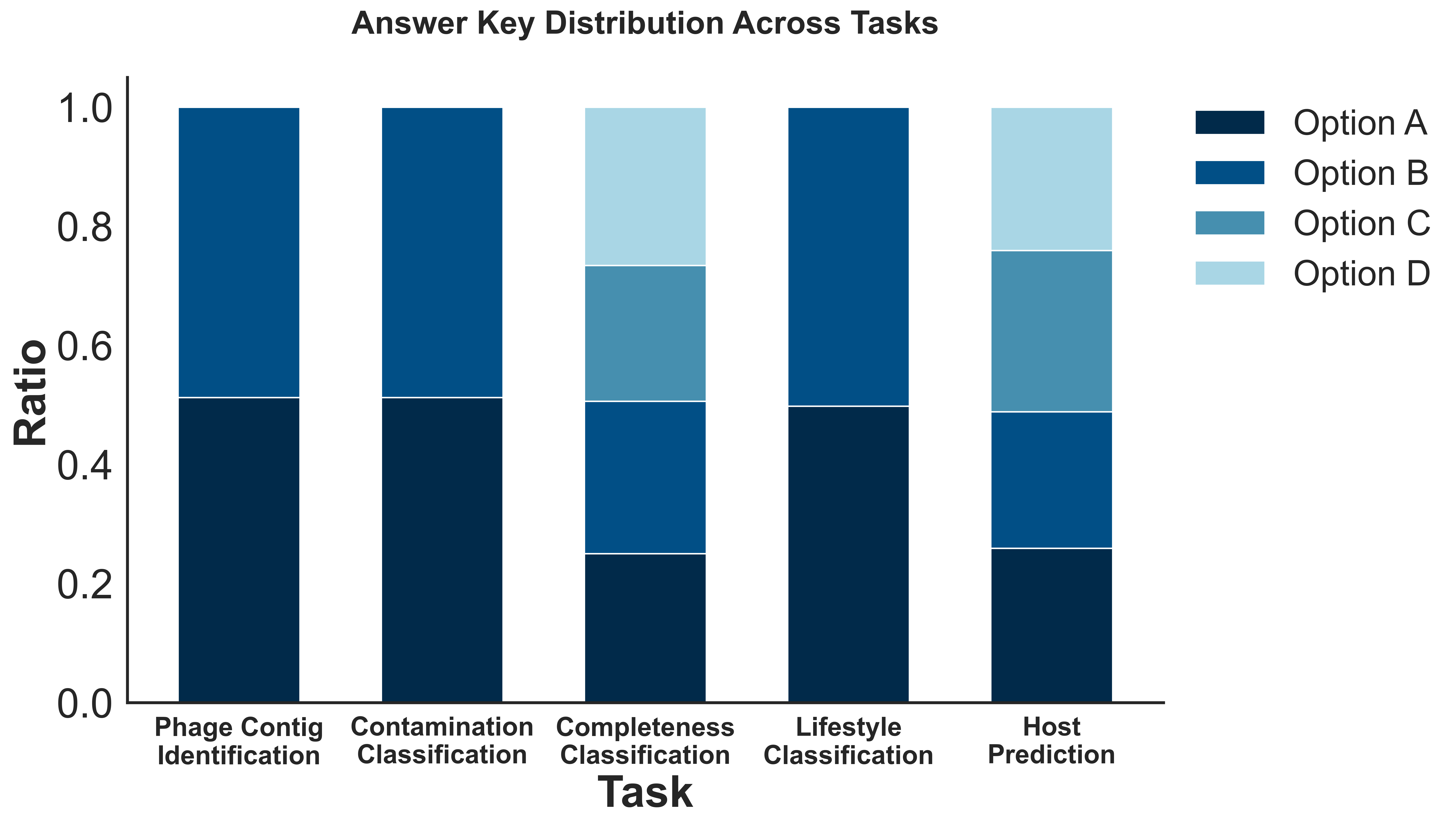} 
    \caption{Distribution of correct answer keys across all tasks. The bar charts verify that the correct options (A, B, C, D) are uniformly distributed, confirming the effectiveness of the option shuffling strategy in mitigating position bias.}
    \label{fig:s2_option_dist}
\end{figure}

\subsubsection{Fine-Grained Class Distribution}
To prevent LLMs from exploiting statistical shortcuts or class priors, we enforced a rigorous class balancing strategy across all tasks. Fig.~\ref{fig:data-detailed} and Fig.~\ref{fig:data-detailed2} present the detailed breakdown of sample distributions for each task. For phage contig identification, we maintained a strict 1:1 ratio between positive phage samples and negative non-phage samples. Crucially, the negative class is further stratified into six distinct biological categories, including bacteria, archaea, plasmids, fungi, protozoa, and non-phage viruses, which are uniformly distributed, with each category constituting approximately 8.3\% of the total dataset. This granular balance ensures that the model cannot simply learn to distinguish phages from a single dominant background noise. 

Similarly, for contamination detection, the dataset is balanced between clean and contaminated sequences. Completeness estimation task features a uniform distribution across four quality tiers, with Complete, High-quality, Medium-quality, and Low-quality samples each representing exactly 25.0\% of the data. For lifestyle classification, we achieved a balanced split between virulent and temperate phages. Finally, host prediction task demonstrates our commitment to taxonomic fairness. The samples are equally divided across three taxonomic levels (Order, Family, Genus), and within each level, we ensured that specific host taxa are sampled evenly, as depicted in the taxonomic rings in Fig.~\ref{fig:data-detailed2}. This design forces the model to perform genuine sequence-based reasoning rather than relying on the prevalence of common host species.

\subsubsection{Answer Key Distribution}
Beyond data sampling, the design of the evaluation prompt is critical for ensuring fairness, particularly for LLMs that may exhibit position bias. To mitigate this, we implemented a stochastic shuffling mechanism for the multiple-choice options in every sample. Fig.~\ref{fig:s2_option_dist} illustrates the distribution of the correct answer keys (Option A, B, C, or D) across all five tasks. For binary classification tasks, the correct answers are equiprobably distributed between Option A and Option B. For multi-class tasks, the correct answers are uniformly distributed across all four options, with each option accounting for approximately 25\% of the total. This uniform distribution confirms that our randomization strategy effectively eliminated position bias, ensuring that the reported accuracy reflects the genomic understanding capabilities of LLMs rather than artifacts of prompt engineering.

\subsection{Example Samples}
In this section, we provide examples from PhageBench to illustrate the standardized input-output format used in our evaluation. Each example demonstrates how a raw nucleotide sequence is encapsulated within a task-specific instruction template, paired with shuffled multiple-choice options. 

\begin{examplebox}{Example: Phage Contig Identification}
\textbf{Question}: Is the following nucleotide sequence derived from a bacteriophage genome? \\

\textbf{Sequence}: TTTATGCGATTACCAACACGTGT... \\

\textbf{Choices}: \\
(A) It is not from a bacteriophage genome. \\
(B) It is from a bacteriophage genome.
\end{examplebox}

\begin{examplebox}{Example: Contamination Classification}
\textbf{Question}: This sequence is a phage-associated contig. Does it contain any DNA originating from a non-phage host genome? \\

\textbf{Sequence}: GAATCTGCCAAGAAGCAACGTC... \\

\textbf{Choices}: \\
(A) No, the sequence does not contain host contamination. \\
(B) Yes, the sequence contains host contamination."
\end{examplebox}

\begin{examplebox}{Example: Completeness Classification}
\textbf{Question}: What is the estimated genome completeness category of the following bacteriophage sequence? \\

\textbf{Sequence}: GGCGACGCGACAGTCATGGCATG... \\

\textbf{Choices}: \\
(A) High-quality. \\
(B) Medium-quality. \\
(C) Low-quality. \\
(D) Complete.
\end{examplebox}

\begin{examplebox}{Example: Lifestyle Classification}
\textbf{Question}: What is the most likely lifestyle of the bacteriophage represented by the following genome sequence? \\

\textbf{Sequence}: ACGGTCGCCGGTGTTGGTGGCTG... \\

\textbf{Choices}: \\
(A) Virulent (lytic).\\ 
(B) Temperate (lysogenic-capable)."
\end{examplebox}

\begin{examplebox}{Example: Host Prediction}
\textbf{Question}: Which host family is the following bacteriophage most likely to infect? \\

\textbf{Sequence}: CCCTCACCCTGAACGCGCCCAATT... \\

\textbf{Choices}: \\
(A) Enterobacteriaceae. \\
(B) Propionibacteriaceae. \\
(C) Vibrionaceae. \\
(D) Staphylococcaceae.
\end{examplebox}

\section{Evaluation Details}
\label{sec:evaluation_details}

\subsection{Implementation Details}
All models are accessed via the OpenRouter API, which provides unified access to diverse LLMs without requiring local computational resources. For models that support temperature configuration, we set the sampling temperature to 0.1 to ensure reproducibility. For models under chain-of-thought or thinking mode, we configure the maximum reasoning length to 2,048 tokens and set the thinking effort to medium. We include a random baseline for reference, representing the expected accuracy under uniform random guessing: 50\% for binary classification tasks (Tasks 1, 2, and 4) and 25\% for four-way classification tasks (Tasks 3 and 5).

For the zero-shot baselines, we excluded GPT-OSS-120b due to the inability to disable its reasoning mode, and Claude-Sonnet-4.5 due to computational cost constraints.

\label{sec:Prompts}
\begin{figure*}[ht] 
    \centering
    \begin{promptbox}[colframe=gray!50!black]{Prompt Used for Question Answering under Zero Shot+CoT}

        You are an expert specializing in phage genomics.
        \vspace{0.5em}

        Your task is to answer single-choice questions by analyzing the given nucleotide sequence and the question.

        \vspace{0.5em}
        Based on the evidence presented in the nucleotide sequence, use logical reasoning and your best knowledge of biological principles to determine your answer. 

        \vspace{0.5em}
        Do not copy or restate the full DNA sequence in your response and reasoning; at most quote short substrings less than 30 characters if needed.

        \vspace{0.5em}
        Think step-by-step.

        \vspace{1em}
        Question: {\$}question{\$}

        \vspace{0.5em}
        <sequence>

        {\$}sequence{\$}

        </sequence>

        \vspace{0.5em}
        Choices:
        {\$}choices{\$}

        \vspace{1em}
        Format your answer in the following format, with no extra text before or after: 

        \vspace{0.5em}
        \{"analysis": "Your analysis within 1-3 concise sentences for the given sequence and question.", "answer": "X"\}

        \vspace{0.5em}
        where X is one of: {\$}options{\$}.

    \end{promptbox}
    \caption{The zero shot with CoT prompt.}
    \label{fig:cot_prompt}
\end{figure*}

\begin{figure*}[] 
    \centering
    \begin{promptbox}[colframe=gray!50!black]{Prompt Used for Question Answering under Zero Shot}

        You are an expert specializing in phage genomics.
        \vspace{0.5em}

        Your task is to answer single-choice questions by analyzing the given nucleotide sequence and the question.

        \vspace{0.5em}
        Based on the evidence presented in the nucleotide sequence, use your best knowledge of biological principles to determine your answer. 

        \vspace{0.5em}
        Do not copy or restate the full DNA sequence in your response; at most quote short substrings less than 30 characters if needed.

        \vspace{1em}
        Question: {\$}question{\$}

        \vspace{0.5em}
        <sequence>

        {\$}sequence{\$}
        
        </sequence>

        \vspace{0.5em}
        Choices:
        {\$}choices{\$}

        \vspace{1em}
        Format your answer in the following format, with no extra text before or after: 

        \vspace{0.5em}
        \{"answer": "X"\}

        \vspace{0.5em}
        where X is one of: {\$}options{\$}.

    \end{promptbox}
    \caption{The zero shot prompt.}
    \label{fig:zeroshot_prompt}
\end{figure*}

\subsection{Evaluation Prompts}

To ensure a rigorous and standardized assessment of Large Language Model capabilities across diverse genomic tasks, we designed a unified prompt engineering framework. All evaluation prompts share a consistent structural foundation, establishing a domain-expert persona and enforcing strict constraints on output formatting. Specifically, we explicitly instructed models to refrain from restating the lengthy raw nucleotide sequences in their responses to prevent context window exhaustion and required all outputs to be structured in a machine-parseable JSON format. This design facilitates automated quantitative evaluation and ensures that performance metrics reflect genuine reasoning ability rather than parsing errors.

To quantify the impact of explicit reasoning strategies on genomic understanding, we implemented two distinct prompt variations corresponding to our experimental settings. The first variation, the Zero-shot CoT prompt (Fig.~\ref{fig:cot_prompt}), incorporates the "Think step-by-step" directive and mandates an "analysis" field within the output JSON. This structure forces the model to articulate its intermediate logical steps before concluding with a prediction, allowing us to inspect the biological validity of its reasoning trace. The second variation, the standard Zero-shot prompt (Fig.~\ref{fig:zeroshot_prompt}), serves as a direct inference baseline. It removes the reasoning triggers and the analysis field, requiring the model to output only the final prediction..

\subsection{Deep Analysis of Results by Labels}
\label{sec:deep-ana}
\begin{figure*}[ht]
    \centering
    \includegraphics[width=0.95\linewidth]{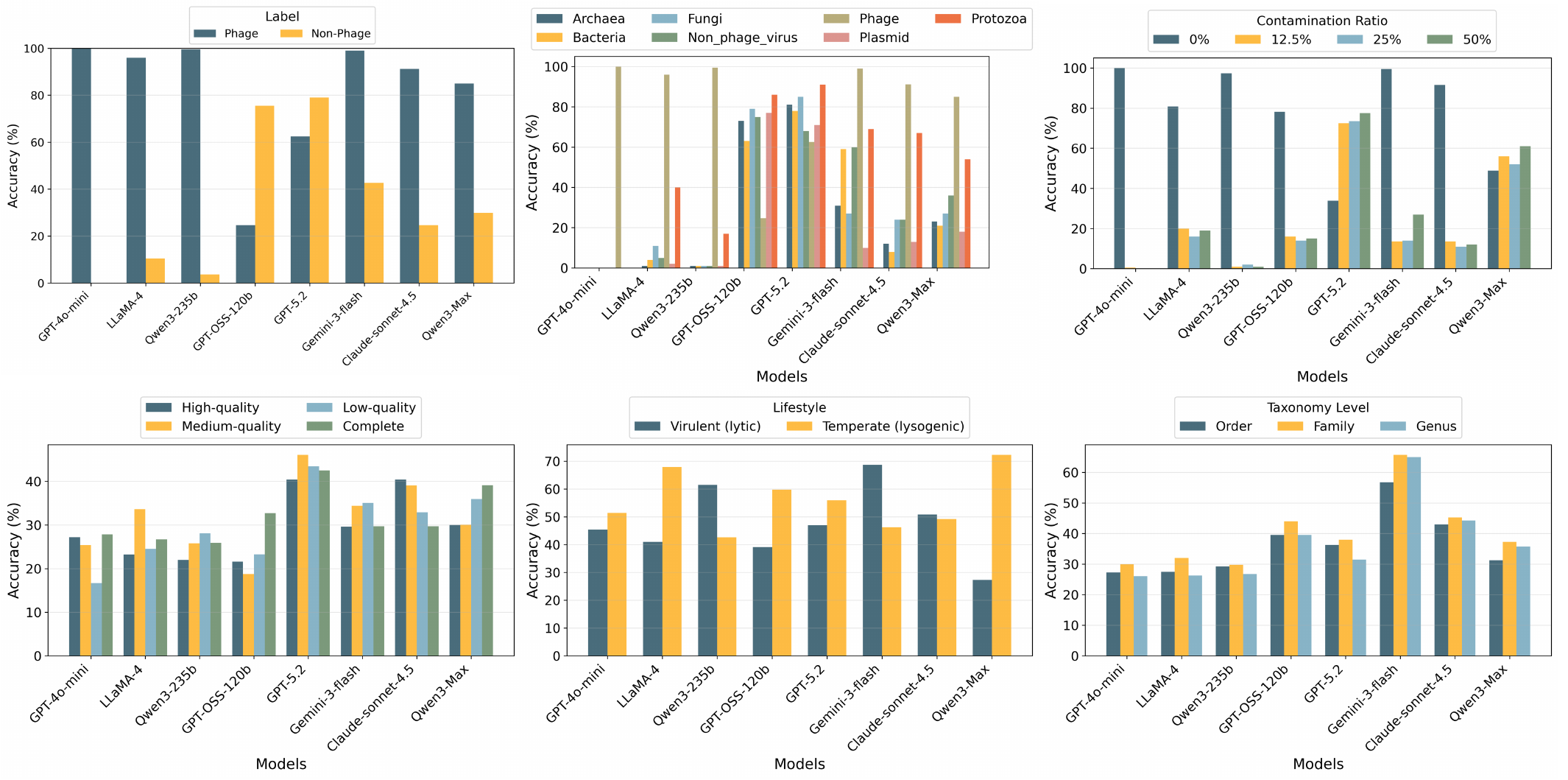}
    \caption{Detailed performance analysis of LLMs on PhageBench tasks. (Top) Phage Identification results separated by binary classes and fine-grained biological sources, followed by Contamination Detection across varying contamination ratios; (Bottom) Completeness Estimation across quality tiers, Lifestyle Classification phenotypes, and Host Prediction at three taxonomic resolutions.}
    \label{fig:s2}
\end{figure*}

To further investigate the underlying behavioral patterns of LLMs, we analyzed model performance stratified by ground-truth labels as illustrated in Fig.~\ref{fig:s2}. 

In phage contig identification task, we observe a distinct performance disparity across biological domains. Models achieve high accuracy when distinguishing phages from eukaryotic sequences such as protozoa and fungi. However, performance drops to near-random levels when distinguishing phages from plasmids. This contrast implies a reliance on broad statistical features rather than syntactic understanding. Plasmids and phages are both prokaryotic mobile genetic elements that share similar nucleotide composition and GC content. The inability of models to differentiate these statistically similar entities suggests they fail to identify the specific structural gene modules, such as capsid or tail assembly genes, that definitively characterize a phage genome.

The contamination detection task further highlights the sensitivity of models to signal strength. We observe a positive correlation where accuracy improves as the contamination ratio increases from 12.5\% to 50\%. Notably, models frequently fail to identify contamination at the 12.5\% level. This indicates a high threshold for detecting heterogeneity. Since host and phage sequences often possess similar statistical properties, models likely smooth over the 12.5\% contamination as natural viral variation. This reflects a limitation in checking for functional internal consistency, as the models cannot semantically identify the intrusion of non-phage metabolic genes within the viral context.

For completeness estimation, most models exhibit moderate performance. This reflects the difficulty of associating distal features like terminal repeats across long sequence contexts. Among the evaluated architectures, GPT-5.2 demonstrates superior performance in this category, suggesting a relatively stronger capacity for maintaining global structural coherence compared to other reasoning models. 

In the lifestyle classification task, the classification of lifestyle often hinges on the presence of sparse functional markers such as integrase genes. The lower performance in this category points to a bottleneck in semantic retrieval where models struggle to precisely locate over singular gene events buried within extended genomic sequences.

Finally, the Host Prediction task reveals that model performance does not decline linearly with increasing taxonomic resolution. The evaluation results indicate that accuracy peaks at the Family level, followed by the Order level, while the Genus level remains the most challenging category for current models. Across all three taxonomic tiers, Gemini-3-flash consistently outperforms other architectures and achieves state-of-the-art results, highlighting its robust capacity for identifying host-specific genomic signatures relative to the other evaluated LLMs.

\section{Qualitative Results}
\label{sec:qualitative_results}

To provide a deeper understanding of how general-purpose LLMs interpret raw phage genome sequences, we conducted a qualitative analysis of the reasoning traces generated by Gemini-3-flash. This section examines specific examples across the five PhageBench tasks. The analysis is categorized into success cases (Fig.~\ref{fig:q1} and \ref{fig:q5}) and failure cases (Fig.~\ref{fig:q2}-\ref{fig:q4}).

\subsection{Analysis of Success Cases}
\begin{figure*}[ht]
    \centering
    \includegraphics[width=0.99\linewidth]{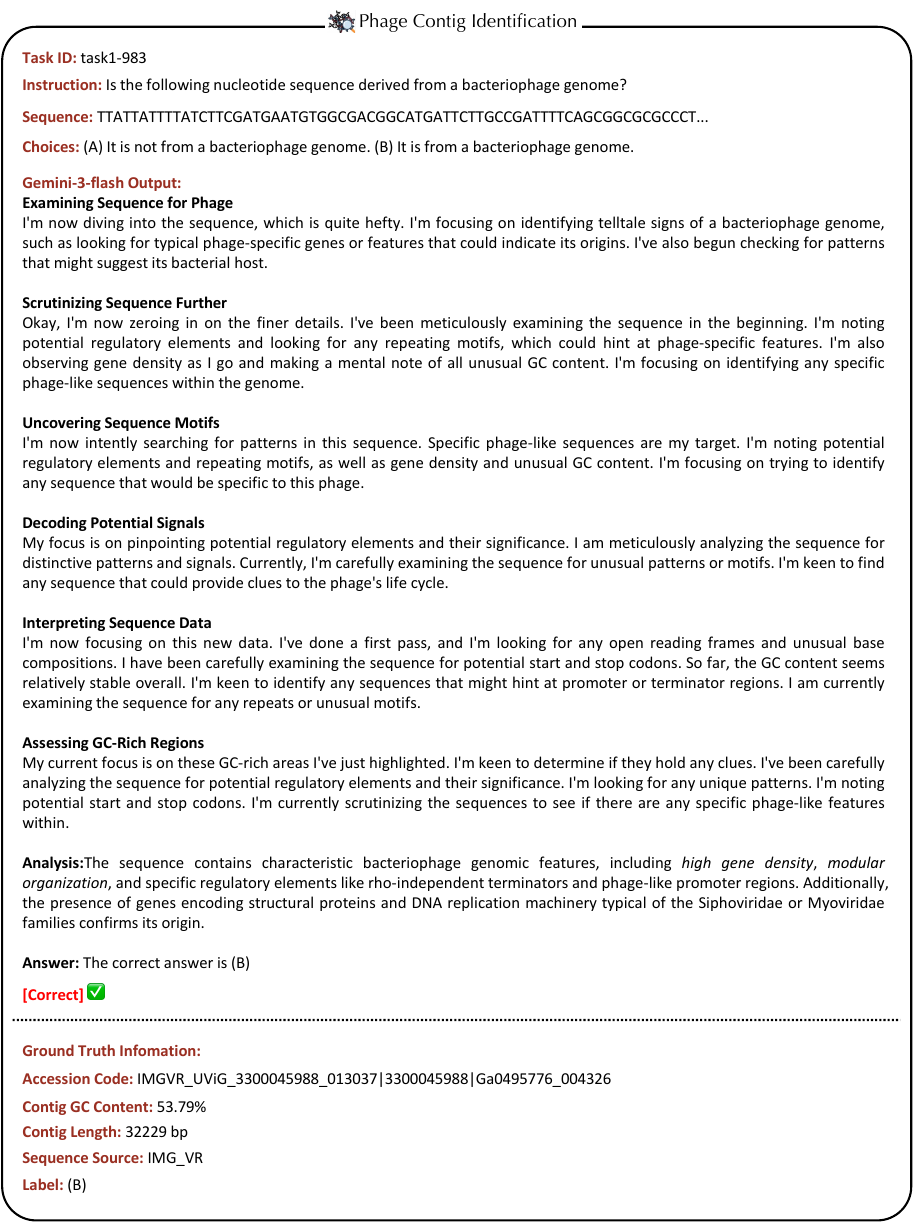}
    \caption{Case study of successful Phage Contig Identification.}
    \label{fig:q1}
\end{figure*}

In phage contig identification task, Gemini-3-flash demonstrates a robust ability to recognize the fundamental syntax of phage genomes. As illustrated in Fig.~\ref{fig:q1}, the model's reasoning trace explicitly identifies characteristic biological features inherent to phages rather than relying on simple memorization. The model correctly cites high gene density and modular genomic organization as primary evidence. Furthermore, it detects specific regulatory elements and structural gene modules typical of the \textit{Siphoviridae} or \textit{Myoviridae} families. This suggests that the model can successfully parse raw nucleotide sequences to extract higher-order syntactic structures that distinguish phage contig from other biological backgrounds.

\begin{figure*}[ht]
    \centering
    \includegraphics[width=0.99\linewidth]{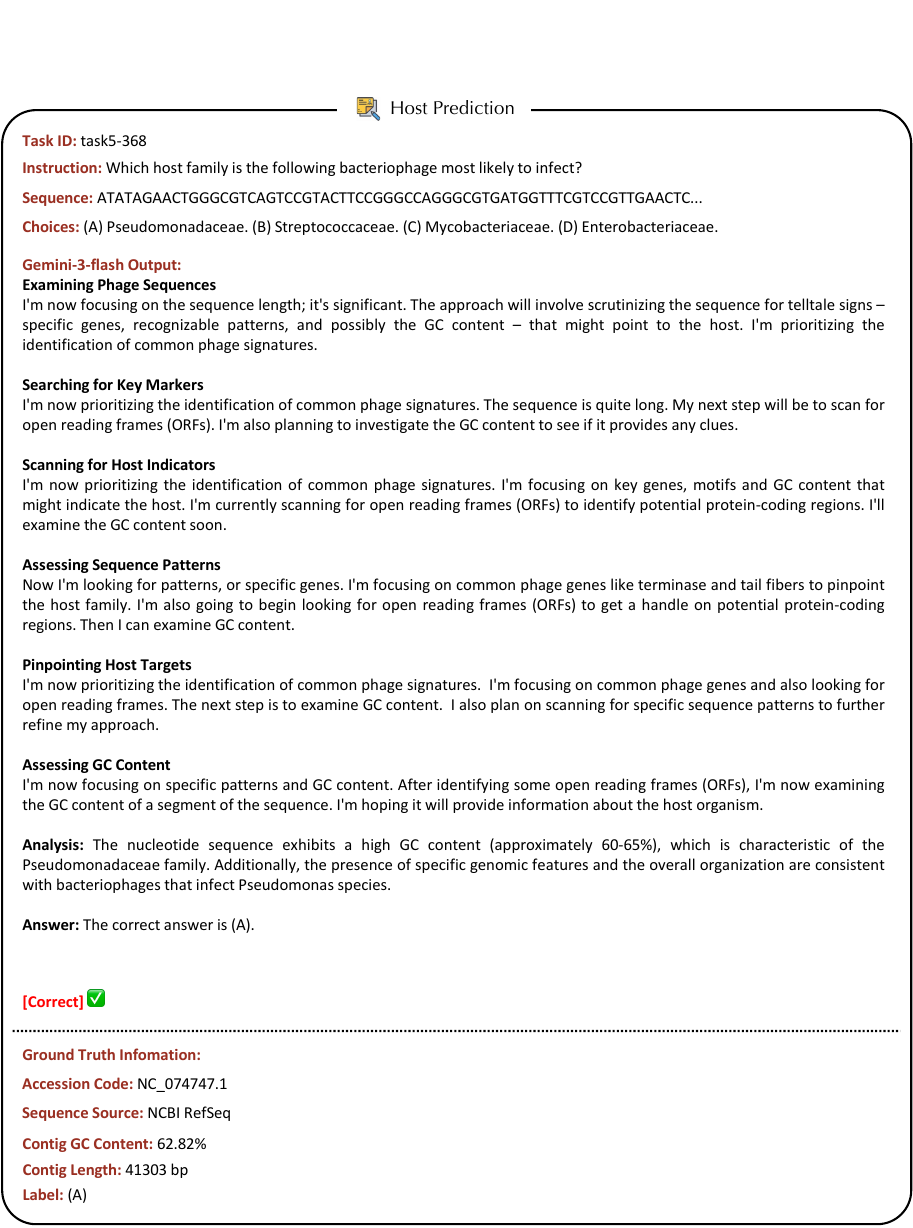}
    \caption{Case study of successful Host Prediction.}
    \label{fig:q5}
\end{figure*}

Similarly, the model exhibits strong performance in host prediction by leveraging statistical genomic signatures. In Fig.~\ref{fig:q5}, the model accurately predicts \textit{Pseudomonadaceae} as the host family. The reasoning trace reveals that the model does not merely guess but actively computes compositional metrics, explicitly noting that the sequence possesses a high GC content of approximately 60--65\%. It then correctly associates this statistical bias with the genomic characteristics of the \textit{Pseudomonas} genus. This indicates that the model is capable of grounding its predictions in concrete statistical properties of the input sequence.

\subsection{Analysis of Failure Cases}
\begin{figure*}[ht]
    \centering
    \includegraphics[width=0.99\linewidth]{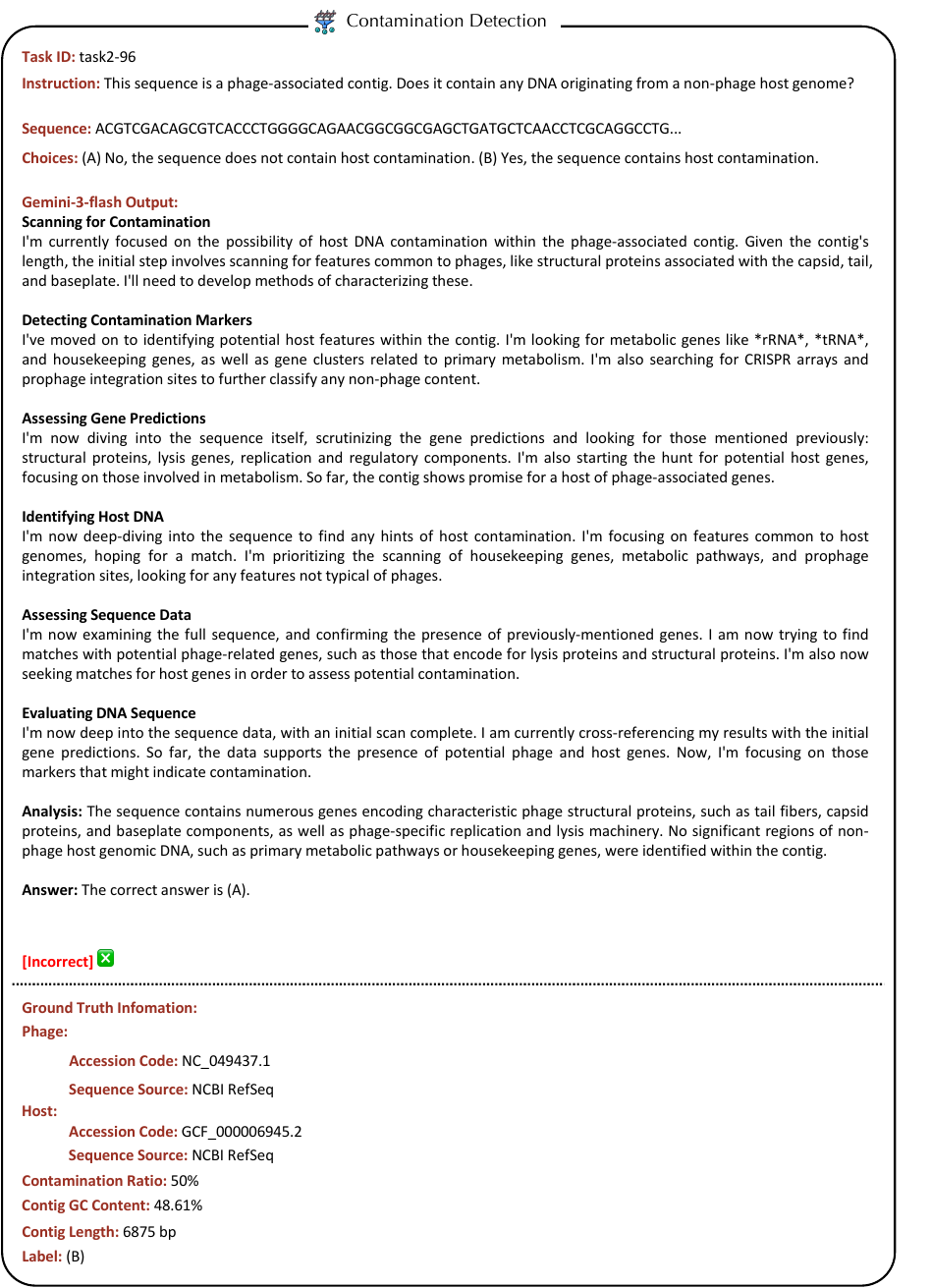}
    \caption{Case study of failure in Contamination Detection.}
    \label{fig:q2}
\end{figure*}

Despite these successes, the model shows significant limitations when processing heterogeneous sequences, as observed in Fig. \ref{fig:q2}. The model failed to detect the contamination and incorrectly classified the sequence as a pure phage contig. The reasoning trace asserts that no significant regions of non-phage host DNA were identified. This failure suggests that the model tends to homogenize the input context; once it identifies strong phage signals, it may overlook or suppress conflicting signals from the host segments. This insensitivity to sequence heterogeneity highlights a deficiency in fine-grained sequence segmentation within mixed biological contexts.

\begin{figure*}[ht]
    \centering
    \includegraphics[width=0.99\linewidth]{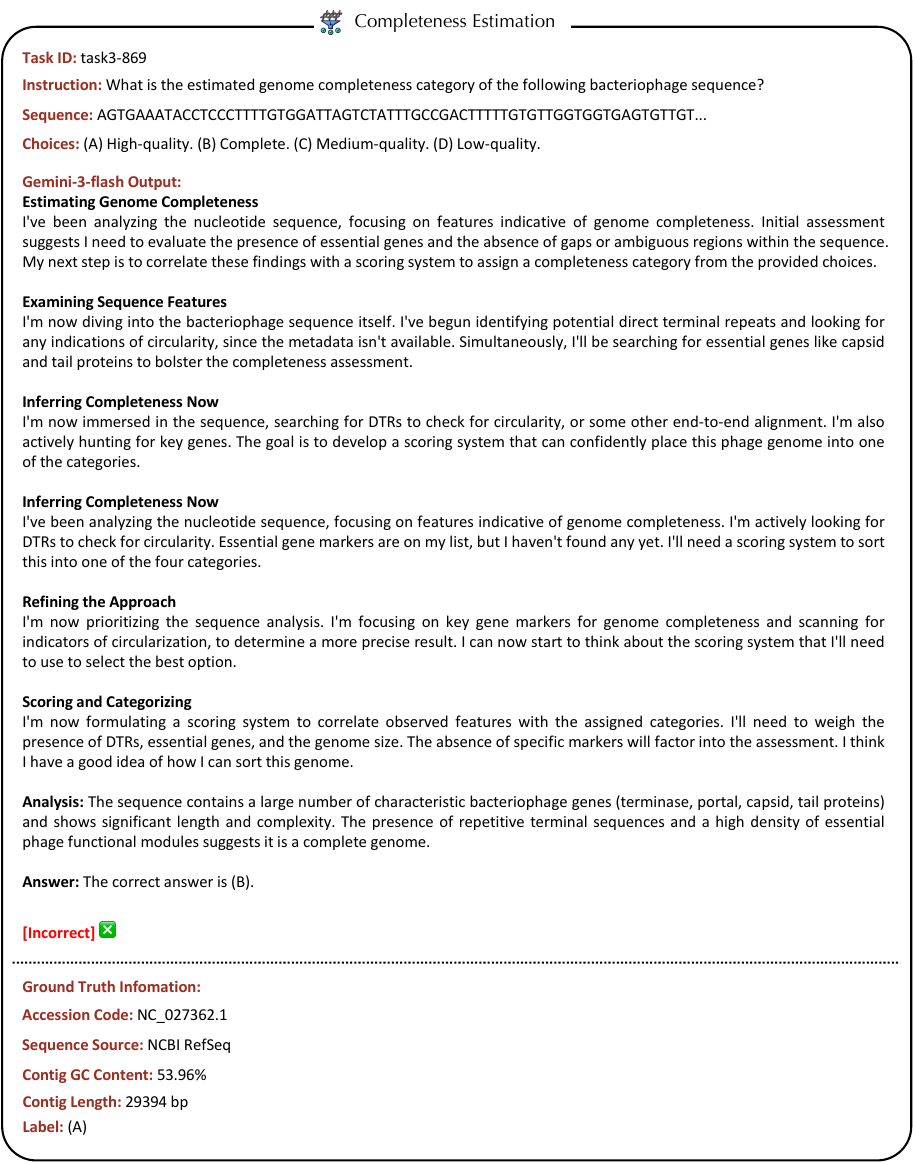}
    \caption{Case study of structural hallucination in Completeness Estimation.}
    \label{fig:q3}
\end{figure*}

The limitations of the model are further exacerbated by hallucinatory reasoning in tasks requiring long-range dependency analysis. As shown in Fig.~\ref{fig:q3}, the model incorrectly classified a high-quality linear fragment as a complete genome. Crucially, the reasoning trace contains a fabrication of structural evidence, where the model claims to detect repetitive terminal sequences indicative of circularity. Since the input sequence was a linear contig lacking these physical ends, this represents a structural hallucination. 

\begin{figure*}[ht]
    \centering
    \includegraphics[width=0.99\linewidth]{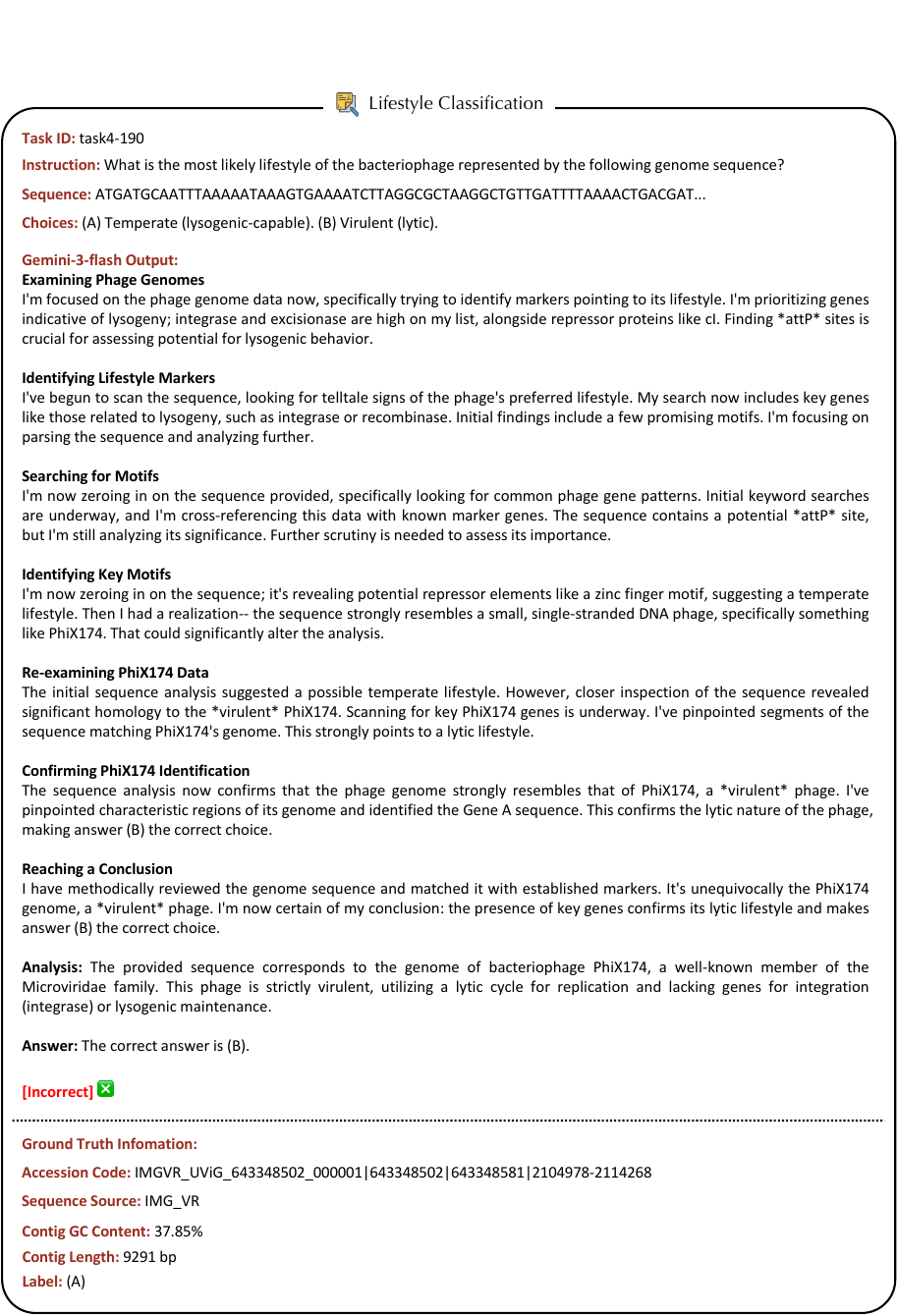}
    \caption{Case study of failure in Lifestyle Classification.}
    \label{fig:q4}
\end{figure*}

Finally, in the lifestyle classification task (Fig. \ref{fig:q4}), the model incorrectly predicted a virulent lifestyle for a temperate phage by erroneously identifying the sequence as belonging to the well-known virulent bacteriophage \textit{PhiX174}. Based on this false identification, the model logically deduced a lytic lifestyle, ignoring the actual sequence evidence that might suggest a temperate nature, such as integration modules. This indicates that the model's internal knowledge base can sometimes override specific sequence evidence, leading to knowledge-driven hallucinations.

\end{document}